# Processing Uncertainty and Indeterminacy in Information Systems projects success mapping

Jose L. Salmeron[a*] and Florentin Smarandache[b]
[a] Pablo de Olavide University at Seville (Spain)
[b] University of New Mexico, Gallup (USA)

ABSTRACT

IS projects success is a complex concept, and its evaluation is complicated, unstructured and not readily quantifiable. Numerous scientific publications address the issue of success in the IS field as well as in other fields. But, little efforts have been done for processing indeterminacy and uncertainty in success research. This paper shows a formal method for mapping success using Neutrosophic Success Map. This is an emerging tool for processing indeterminacy and uncertainty in success research. EIS success have been analyzed using this tool.

1. Introduction

For academics and practitioners concerned with computer-based Information Systems (IS), one central issue is the study of development and implementation project success. Literature (Barros et al., 2004; Poon and Wagner, 2001; Rainer and Watson, 1995; Redmil, 1990) suggest that IS projects have lower success rates than other technical projects. Irrespective of the accuracy of this presumption, the number of unsuccessful IS projects are over the number of successful ones. Therefore, it is worthwhile to develop a formal method for mapping success, since proper comprehension of the complex nature of IS success is critical for the successful application of technical principles to this discipline.

---

* Corresponding author
  e-mail address: jlsalsil@upo.es  (J.L. Salmeron)

To increase the chances of an IS project to be perceived as successful for people involved in project, it is necessary to identify at the outset of the project what factors are important and influencing that success. These are the Critical Success Factors (CSF) of the project. Whereas several CSF analyses appear in the literature, most of them do not have any technical background. In addition, almost none of them focus on relations between them. In addition, it is important to discover the relationships between them. Research about it was becoming scarce.

In this paper, we propose the use of an innovative technique for processing uncertainty and indeterminacy to set success maps in IS projects. The main strengths of this paper are two-folds: it provides a method for processing indeterminacy and uncertainty within success and it also allows building a success map.

The remainder of this paper is structured as follows: Section 2 shows previous research; Section 3 reviews cognitive maps and its evolution; Section 4 is focused on the research model; Section 5 presents and analyzes the results; the final section shows the paper's conclusions.

2. Previous research

Success is not depending to just one issue. Complex relations of interdependence exist between IS, organization, and users. Thus, for example, reducing costs in an organization cannot be derived solely from IS implementation. Studies indicate that the IS success is hard to assess because it represent a vague topic that does not easily lend itself to direct measurement (DeLone and McLean, 1992).

According to Zviran and Erlich (2003), academics tried to assess the IS success as a function of cost-benefit (King and Schrems, 1978), information value (Epstein and King,

1983; Gallagher, 1974), or organization performance (Turner, 1982). System acceptance (Davis, 1989) has used for it too. Anyway, cost-benefit, information value, system acceptance, and organization performance are difficult to apply as measures.

Critical Success Factor method (Rockart, 1979) have been used as a mean for identifying the important elements of IS success since 1979. It was developed as a method to enable CEOs to recognize their own information needs so that IS could be built to meet those needs. This concept has received a wide diffusion among IS scholars and practitioners (Butler and Fitzgerald, 1999).

Numerous scientific publications address the issue of CSF in the IS field as well as in other fields. But, little efforts have been done for introducing formal methods in success research. Some authors (Poon and Wagner, 2001) analysed some aspects of CSF just by the use of personal interviews, whereas others (Ragahunathan et al., 1989) carried out a Survey-based field study. Interviews and/or questionnaires are common tools for measuring success. However, formal methodology is not usual.

On the other hand, Salmeron and Herrero (2005) propose a hierarchical model to model success. Anyway, indeterminacy was not processed. Therefore, we think that a formal method to process indeterminacy and uncertainty in IS success is an useful endeavour.

3. Uncertainty and Indeterminacy processing in cognitive maps

3.1. Cognitive mapping

A cognitive map shows a representation of how humans think about a particular issue, by analyzing, arranging the problems and graphically mapping concepts that are connected between them. In addition, it identifies causes and effects and explains causal links (Eden and Ackermann, 1992). The cognitive maps study perceptions about the world and the way they act to reach human desires within their world. Kelly (1955, 1970) gives the foundation for this theory, based on a particular cognitive psychological body of knowledge. The base postulate for the theory is that "a person's processes are psychologically canalized by the ways in which he anticipates events." Mental models of top managers in firms operating in a competitive environment have been studied (Barr et al., 1992) using cognitive mapping. They suggest that the cognitive models of these managers must take into account significant new areas of opportunity or technological developments, if they want stay ahead. In this sense, it is critical to consider mental models in success research.

3.2. Neutrosophic Cognitive Maps (NCM)

In fact, success is a complex concept, and its evaluation is complicated, unstructured and not readily quantifiable. The NCM model seems to be a good choice to deal with this ambiguity. NCM are flexible and can be customised in order to consider the CSFs of different IT projects.

Neutrosophic Cognitive Maps (Vasantha-Kandasamy and Smarandache, 2003) is based on Neutrosophic Logic (Smarandache, 1999) and Fuzzy Cognitive Maps. Neutrosophic Logic emerges as an alternative to the existing logics and it represents a mathematical model of uncertainty, and indeterminacy. A logic in which each proposition is estimated to have the percentage of truth in a subset T, the percentage of indeterminacy in a subset I, and the percentage of falsity in a subset F, is called Neutrosophic Logic. It uses a subset of truth (or indeterminacy, or falsity), instead of using a number, because in many cases, humans are not able to exactly determine the percentages of truth and of falsity but to approximate them: for example a proposition is between 30-40% true. The subsets are not necessarily intervals, but any sets (discrete, continuous, open or closed or half-open/ half-closed interval, intersections or unions of the previous sets, etc.) in accordance with the given proposition. A subset may have one element only in special cases of this logic. It is imperative to mention here that the Neutrosophic logic is a strait generalization of the theory of Intuitionistic Fuzzy Logic.

Neutrosophic Logic which is an extension/combination of the fuzzy logic in which indeterminacy is included. It has become very essential that the notion of neutrosophic logic play a vital role in several of the real world problems like law, medicine, industry, finance, IT, stocks and share, and so on. Fuzzy theory measures the grade of membership or the non-existence of a membership in the revolutionary way but fuzzy theory has failed to attribute the concept when the relations between notions or nodes or concepts in problems are indeterminate. In fact one can say the inclusion of the concept of indeterminate situation with fuzzy concepts will form the neutrosophic concepts (there also is the neutrosophic set, neutrosophic probability and statistics).

In this sense, Fuzzy Cognitive Maps mainly deal with the relation / non-relation between two nodes or concepts but it fails to deal with the relation between two conceptual nodes when the relation is an indeterminate one. Neutrosophic logic is the only tool known to us, which deals with the notions of indeterminacy.

A Neutrosophic Cognitive Map (NCM) is a neutrosophic directed graph with concepts like policies, events, etc. as nodes and causalities or indeterminates as edges. It represents the causal relationship between concepts. A neutrosophic directed graph is a directed graph in which at least one edge is an indeterminacy denoted by dotted lines.

Let $C_1, C_2,\ldots, C_n$ denote n nodes, further we assume each node is a neutrosophic vector from neutrosophic vector space V. So a node $C_i$ will be represented by $(x_1, \ldots, x_n)$ where $x_k$'s are zero or one or I (I is the indeterminate introduced before) and $x_k = 1$ means that the node $C_k$ is in the on state and $x_k = 0$ means the node is in the off state and $x_k = I$ means the nodes state is an indeterminate at that time or in that situation.

Let $C_i$ and $C_j$ denote the two nodes of the NCM. The directed edge from $C_i$ to $C_j$ denotes the causality of $C_i$ on $C_j$ called connections. Every edge in the NCM is weighted with a number in the set {-1, 0, 1, I}. Let $e_{ij}$ be the weight of the directed edge $C_iC_j$, $e_{ij} \in \{-1,0,1,I\}$. $e_{ij} = 0$ if $C_i$ does not have any effect on $C_j$, $e_{ij} = 1$ if increase (or decrease) in $C_i$ causes increase (or decreases) in $C_j$, $e_{ij} = -1$ if increase (or decrease) in $C_i$ causes decrease (or increase) in $C_j$. $e_{ij} = I$ if the relation or effect of $C_i$ on $C_j$ is an indeterminate.

The edge $e_{ij}$ takes values in the fuzzy causal interval [–1, 1] ($e_{ij} = 0$ indicates no causality, $e_{ij} > 0$ indicates causal increase; that $C_j$ increases as $C_i$ increases and $C_j$ decreases as $C_i$ decreases, $e_{ij} < 0$ indicates causal decrease or negative causality $C_j$ decreases as $C_i$ increases or $C_j$, increases as $C_i$ decreases. Simple FCMs have edge value in {-1, 0, 1}. Thus if causality occurs it occurs to maximal positive or negative degree.

It is important to note that $e_{ij}$ measures only absence or presence of influence of the node $C_i$ on $C_j$ but till now any researcher has not contemplated the indeterminacy of any relation between two nodes $C_i$ and $C_j$. When we deal with unsupervised data, there are situations when no relation can be determined between some two nodes. In our view this will certainly give a more appropriate result and also caution the user about the risks and opportunities of indeterminacy.

Using NCM is possible to build a Neutrosophic Success Map (NSM). NSM nodes represent Critical Success Factors (CSF). They are the limited number of areas in which results, if they are satisfactory, will ensure successful competitive performance for the organization. They are the few key areas where "things must go right" for the project (Rockart, 1979). This tool shows the relations and the fuzzy values within in an easy understanding way. This is an useful approach for non-technical decision makers. At the same time, it allows computation as FCM. Figure 1 shows the NSM static context.

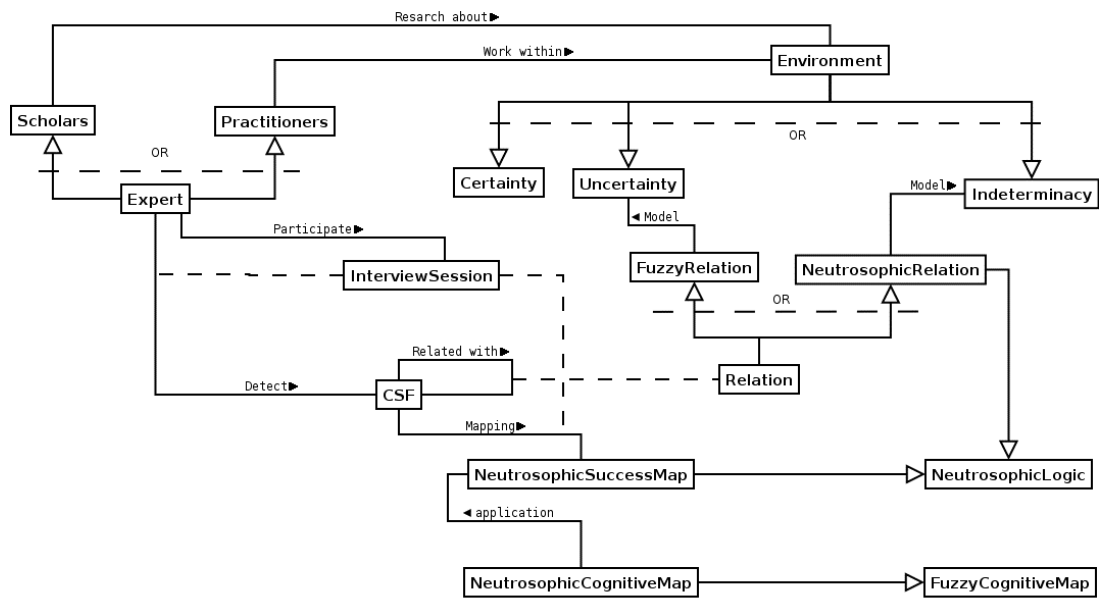

Figure 1: NSM static context

## 5. Building a NSM

EIS project have been used for building a Neutrosophic Success Map. EIS, or executive support systems as they are sometimes called, can be defined as computer-based information systems that support communications, coordination, planning and control functions of managers and executives in organizations (Salmeron and Herrero, 2005).

NSM will be based on textual descriptions given by EIS experts on interviews with them. The steps followed are:

1) Experts selection. It is critical step. Expert selection was based on specific knowledge of EIS systems. Experts are 19 EIS users of leading companies and EIS researchers. The composition of the respondents is important. Multiple choices were contemplated. The main selection criterion considered was recognized knowledge in research topic, absence of conflicts of interest and geographic diversity. All conditions were respected. In addition, respondents were not chosen just because they are easily accessible.
2) Identification of CSF influencing the EIS systems.
3) Identification and assess of causal relationships among these CSF. Indeterminacy relations are included.

Experts discover the CSFs and give qualitative estimates of the strengths associated with causal links between nodes representing these CSFs. These estimates, often expressed in imprecise or fuzzy/neutrosophic linguistic terms, are translated into numeric values in the range −1 to 1. In addition, indeterminacy is used for modelling that kind of relations relationships among nodes.

The nodes (CSFs) discover was the following:

1. Users' involvement ($x_1$). It is defined as a mental or psychological state of users toward the system and its development process. It is generally accepted that IS users' involvement in the application design and implementation is important and necessary (Hwang and Thorn, 1999). It is essential in maintenance phase too.
2. Speedy prototype development ($x_2$). It encourages the right information needs because it interacts between user and system as soon as possible.

3. Top management support ($x_3$). EIS support with his/her authority and influence over the rest of the executives.
4. Flexible system ($x_4$). EIS must be flexible enough to be able to get adapted to changes in the types of problems and the needs of information.
5. Right information requirements ($x_5$). Eliciting requirements is one of the most complicated tasks in developing EIS and getting a correct requirement set is challenging.
6. Technological integration ($x_6$). EIS tool selected must be integrated in companies' technological environment.
7. Balanced development team ($x_7$). Suitable human resources are required for developing EIS. Technical background and business knowledge are needed.
8. Business value ($x_8$). The system must solve a critical business problem. There should be a clear business value in EIS use.
9. Change management ($x_9$). It is the process of developing a planned approach to change in a firm. EIS will be a new way of working. Typically the objective is to maximize the collective efforts of all people involved in the change and minimize the risk of failure of EIS project.

The NSM find out is presented in Figure 2. Fuzzy values are included.

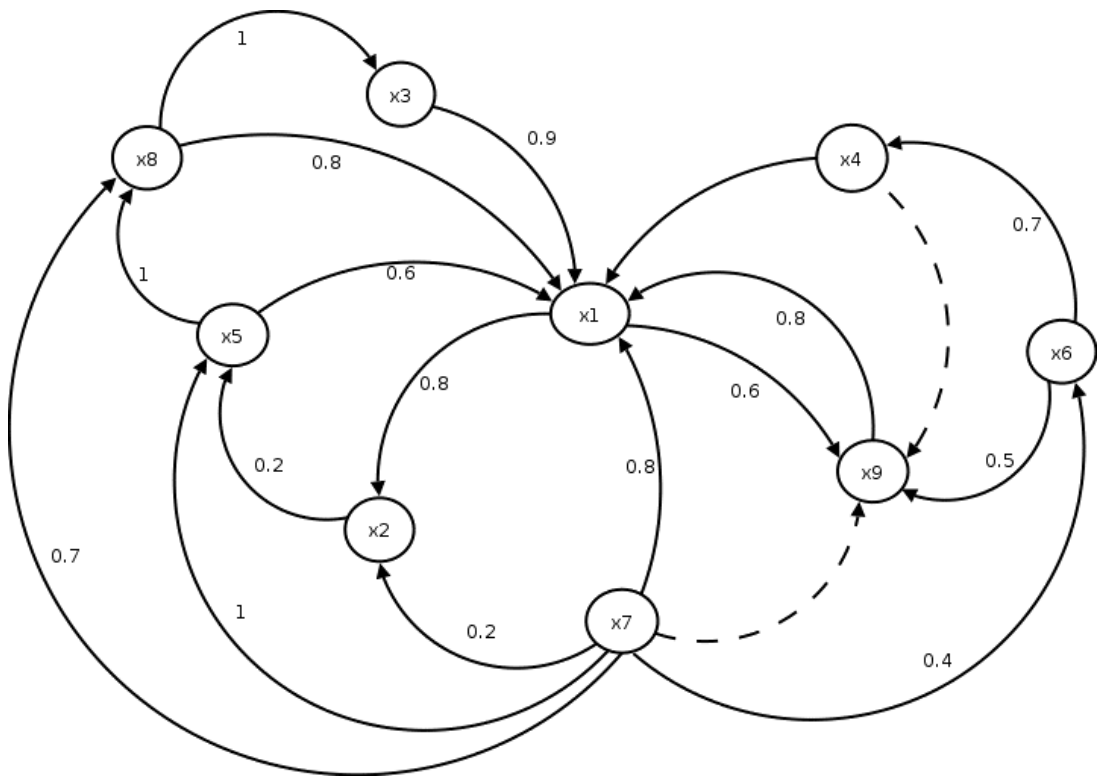

Figure 2. EIS NSM

The adjacency matrix associated to NSM is N(E).

$$N(E) = \begin{pmatrix} 0 & 0.8 & 0 & 0 & 0 & 0 & 0 & 0 & 0.6 \\ 0 & 0 & 0 & 0 & 0.2 & 0 & 0 & 0 & 0 \\ 0.9 & 0 & 0 & 0 & 0 & 0 & 0 & 0 & 0 \\ 0.1 & 0 & 0 & 0 & 0 & 0 & 0 & 0 & I \\ 0.6 & 0 & 0 & 0 & 0 & 0 & 0 & 1 & 0 \\ 0 & 0 & 0 & 0.7 & 0 & 0 & 0 & 0 & 0.5 \\ 0.8 & 0.2 & 0 & 0 & 1 & 0.4 & 0 & 0.7 & I \\ 0.8 & 0 & 1 & 0 & 0 & 0 & 0 & 0 & 0 \\ 0.8 & 0 & 0 & 0 & 0 & 0 & 0 & 0 & 0 \end{pmatrix}$$

The stronger relations are between $x_5$ to $x_8$, $x_7$ to $x_5$ and $x_8$ to $x_3$. It follows that a balanced development team has a positive influence over elicitation requirements process. In the same sense, eliciting right requirements have a positive influence over system business value and system business value over top management support. In addition, users' involvement receives influence from six nodes.

On the other hand, we have found two neutrosophic relations between $x_4$ to $x_9$, $x_5$ to $x_3$ and $x_7$ to $x_9$. It follows that experts perceive indeterminacy in relations between EIS flexibility and balanced team skills to change management. They can not to assess the relation between them, but they perceive that relation could exist. It is an useful information since the decision-makers can be advised from it. They will be able to be careful with those relations.

In addition, NSM predict effects of one or more CSFs (nodes) in the regarding ones. If we know that any CSFs are on, we can discover the influence over the others. This process is similar in Fuzzy Cognitive Maps.

Let $\overrightarrow{C_1C_2}, \overrightarrow{C_2C_3},..., \overrightarrow{C_{n-1}C_n}$ be cycle (Vasantha-Kandasamy and Smarandache, 2003), when $C_i$ is switched on and if the causality flow through the edges of a cycle and if it again causes $C_i$, we say that the dynamical system goes round and round. This is true for any node $C_i$, for i = 1, 2,..., n. the equilibrium state for this dynamical system is called the hidden pattern. If the equilibrium state of a dynamical system is a unique state vector, then it is called a fixed point. If the NSM settles with a state vector repeating in the form

$$x_1 \to x_2 \to ... \to x_i \to x_1,$$

then this equilibrium is called a limit cycle of the NSM.

Let $C_1, C_2,..., C_n$ be the CSFs of an NSM. Let E be the associated adjacency matrix. Let us find the hidden pattern when $x_1$ is switched on when an input is given as the vector $A_1$ = (1, 0, 0,..., 0), the data should pass through the neutrosophic matrix N(E), this is done by multiplying $A_1$ by the matrix N(E). Let $A_1 N(E) = (a_1, a_2,..., a_n)$ with the threshold operation that is by replacing $a_i$ by 1 if $a_i > k$ and $a_i$ by 0 if $a_i < k$ and $a_i$ by I if $a_i$ is not a integer.

$$f(k) = \begin{cases} a_i < k \Rightarrow a_i = 0 \\ a_i > k \Rightarrow a_i = 1 \\ a_i = b + c \times I \Rightarrow a_i = b \\ a_i = c \times I \Rightarrow a_i = I \end{cases}$$

This procedure is repeated till we get a limit cycle or a fixed point. According to this, the limit cycle or a fixed point of vector state of each CSFs is calculated with k=0.5. We take the state vector $A_1$ = (1 0 0 0 0 0 0). We will see the effect of $A_1$ over the model.

$$A_1 N(E) = (0 \quad 0.8 \quad 0 \quad 0 \quad 0 \quad 0 \quad 0 \quad 0 \quad 0.6) \longrightarrow (1 \quad 1 \quad 0 \quad 0 \quad 0 \quad 0 \quad 0 \quad 0 \quad 1) = A_2$$

$$A_2 N(E) = (0.8 \quad 0.8 \quad 0 \quad 0 \quad 0.2 \quad 0 \quad 0 \quad 0 \quad 0.6) \longrightarrow (1 \quad 1 \quad 0 \quad 0 \quad 0 \quad 0 \quad 0 \quad 0 \quad 1) = A_3 = A_2$$

$A_2$ is a fixed point. According with experts the on state of users' involvement has effect over speedy prototype development and change management.

We take the new state vector $A_1$ = (1 0 1 0 0 0 0 0 0). We will see the effect of users' involvement and top management support ($A_1$) over the model.

$$A_1 N(E) = (0.9 \quad 0.8 \quad 0 \quad 0 \quad 0.2 \quad 0 \quad 0 \quad 0 \quad 0.6) \longrightarrow (1 \quad 1 \quad 1 \quad 0 \quad 0 \quad 0 \quad 0 \quad 0 \quad 1) = A_2$$

$$A_2 N(E) = (1.7 \quad 0.8 \quad 0 \quad 0 \quad 0.2 \quad 0 \quad 0 \quad 0 \quad 0.6) \longrightarrow (1 \quad 1 \quad 1 \quad 0 \quad 0 \quad 0 \quad 0 \quad 0 \quad 1) = A_3 = A_2$$

Thus $A_2$=(1 1 1 0 0 0 0 0 1), according with experts the on state of users' involvement and top management support have effects over the prototype speed of development ($x_2$) and change management ($x_9$). It is interesting to discover that both previous state vectors have the same influence over the model. Both vector states have influence over prototype speed of development ($x_2$) and change management, but no direct effect over the rest of CSFs.

The vector states described are only two of the several available, even vectors with several CSFs on. However the proposal here presented is as simple as possible while being consistent with the process, data gathered from the expert's perceptions, and the aims and objectives of the paper.

6. Conclusions

The main strengths of this paper are two-folds: it provides a method for project success mapping and it also allows know CSF effects over the other ones. In this paper, we proposed the use of the Neutrosophic Success Maps to map EIS success.

A tool for evaluating suitable success models for IS projects is required due to the increased complexity and uncertainty associated to this kind of projects. This leads to the innovative idea of adapting and improving the existent Neutrosophic theories for their application to indicators of success for IS projects.

Neutrosophic Success Map is an innovative success research approach. NSM is based on Neutrosophic Cognitive Map. The concept of NCM can be used in modelling of systems success, since the concept of indeterminacy play a role in that topic. This was our main aim is to use NCMs in place of FCMs. When an indeterminate causality is present in an FCM we term it as an NCM.

The results not mean that any CSF is unimportant or has not effect over the model. It means what are the respondents' perceptions about the relationships of them. This is a main issue, since it is possible to manage the development process with more information about the expectations of final users.

Anyway, more research is needed about Neutrosophic logic limit and applications. Incorporating the analysis of NCM and NSM, the study proposes an innovative way for success research. We think this is an useful endeavour.